\documentclass[10pt,twocolumn,letterpaper]{article}

\usepackage{cvpr}              

\usepackage[accsupp]{axessibility}
\usepackage{graphicx}
\usepackage{amsmath}
\usepackage{amssymb}
\usepackage{booktabs}
\usepackage{newtxmath}
\usepackage{bm}
\usepackage{algorithm}
\usepackage{algorithmic}
\usepackage{xcolor}

\usepackage[pagebackref,breaklinks,colorlinks]{hyperref}

\usepackage[capitalize]{cleveref}
\crefname{section}{Sec.}{Secs.}
\Crefname{section}{Section}{Sections}
\Crefname{table}{Table}{Tables}
\crefname{table}{Tab.}{Tabs.}

\def\cvprPaperID{4485} 
\def\confName{CVPR}
\def\confYear{2023}

\newcommand\tabcaption{\def\@captype{table}\caption}
\newcommand\figcaption{\def\@captype{figure}\caption}

\begin{document}

\title{No One Left Behind: Improving the Worst Categories in Long-Tailed Learning}

\author{
Yingxiao Du,~Jianxin Wu\thanks{J. Wu is the corresponding author. This research was partly supported by the National Natural Science Foundation of China under Grant 62276123 and Grant 61921006.}\\
State Key Laboratory for Novel Software Technology, Nanjing University\\
Nanjing, China, 210023 \\
{\tt\small duyx@lamda.nju.edu.cn,~wujx2001@gmail.com}
}
\maketitle

\begin{abstract}
Unlike the case when using a balanced training dataset, the per-class recall (i.e., accuracy) of neural networks trained with an imbalanced dataset are known to vary a lot from category to category. The convention in long-tailed recognition is to manually split all categories into three subsets and report the average accuracy within each subset. We argue that under such an evaluation setting, some categories are inevitably sacrificed. On one hand, focusing on the average accuracy on a balanced test set incurs little penalty even if some worst performing categories have zero accuracy. On the other hand, classes in the ``Few'' subset do not necessarily perform worse than those in the ``Many'' or ``Medium'' subsets. We therefore advocate to focus more on improving the lowest recall among all categories and the harmonic mean of all recall values. Specifically, we propose a simple plug-in method that is applicable to a wide range of methods. By simply re-training the classifier of an existing pre-trained model with our proposed loss function and using an optional ensemble trick that combines the predictions of the two classifiers, we achieve a more uniform distribution of recall values across categories, which leads to a higher harmonic mean accuracy while the (arithmetic) average accuracy is still high. The effectiveness of our method is justified on widely used benchmark datasets.
\end{abstract}

\section{Introduction}\label{sec:introduction}

Various gaps exist when adapting image recognition techniques that are developed in the lab to industrial applications. The most noteworthy one is perhaps the difference between training datasets. Most training datasets used in academic research~\cite{deng2009imagenet,krizhevsky2009cifar} are balanced with respect to the number of images per class. This should not be taken for granted because datasets used in real-world applications are more likely to be imbalanced. Training deep models on these datasets is not trivial as models are known to perform poorly on such datasets. Long-tailed recognition is a research field that aims at tackling this challenge.

\begin{figure}
    \centering
    \includegraphics[width=\linewidth]{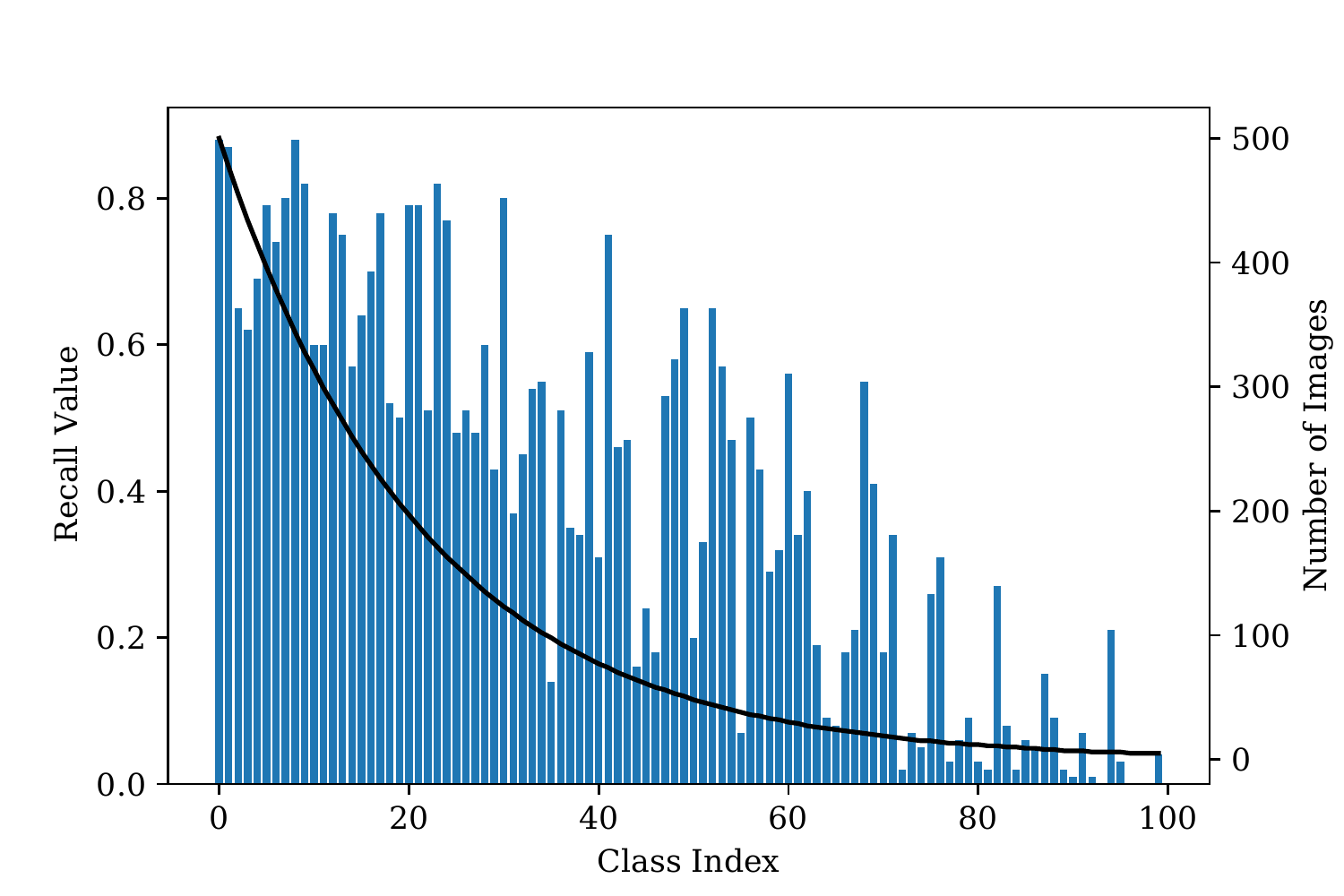}
    \caption{The per-class recall of models trained on the imbalanced CIFAR100 (with imbalance ratio 100). Per-class recall value varies a lot from category to category. Moreover, it is not necessarily true that all categories in the ``Few'' subset have lower accuracy than those in the ``Many'' or ``Medium'' subsets.}
    \label{fig:per-class-recall}
\end{figure}

\begin{table}
    \centering
	\small
    \begin{tabular}{lcc}
    \toprule
    Method&Mean Accuracy&Lowest Recall\\
    \midrule
    BSCE~\cite{ren2020bsce}&42.24&3.00\\
    DiVE~\cite{he2021dive}&45.11&2.00\\
    MiSLAS~\cite{zhong2021improving-calibration}&47.05&5.00\\
    RIDE~\cite{wang2021ride}&48.64&2.00\\
    PaCo~\cite{cui2021paco}&51.24&5.00\\
    \bottomrule
    \end{tabular}
    \tabcaption{The lowest per-class recall of various state-of-the-art methods on the imbalanced CIFAR100 (with imbalance ratio 100). Although there are rapid improvements over the mean accuracy, the lowest per-class recall remains very low.}
    \label{tab:lowest-recall}
\end{table}

By plotting the per-class recall (i.e., per-class accuracy) of the model trained on the imbalanced CIFAR00 dataset (with imbalance ratio 100) in~\cref{fig:per-class-recall}, we find that the recall varies dramatically from category to category. We argue that in real world applications, \emph{all categories are equally important and no one should be left behind}. We also find that despite the rapid developments in this community, no obvious improvement over the lowest per-class recall is witnessed in the past few years~\cite{ren2020bsce,he2021dive,zhong2021improving-calibration,wang2021ride,cui2021paco}, as is shown in~\cref{tab:lowest-recall}. The convention in long-tailed recognition research is to split the classes into three subsets based on the number of training images. The accuracy within each subset is often reported along with the overall accuracy. While this evaluation scheme seems reasonable at the first glance, we argue it is potentially problematic. First of all, computing the average recall within each subset is way too coarse, making it impossible to reflect whether some classes are completely sacrificed but covered up by other ``easy'' tail classes. What's more, we find it is not necessarily true that classes in the ``Few'' category all have lower recall than classes in the other two subsets, as is shown in~\cref{fig:per-class-recall}. Therefore, focusing on improving the mean accuracy alone is not enough, especially in real-world applications.

In this paper, we propose a novel method to make sure that no category is left behind. Specifically, we argue that although mean accuracy is widely used in image classification as an optimization objective, due to the fact that different classes have very different recall in long-tailed recognition, it is not the most suitable objective as it incurs little penalty even if some categories have very small per-class accuracy values (e.g., close to 0). Hence, to improve even the worst-performing categories, we believe the harmonic mean of per-class recall would be a better objective. Since harmonic mean is very sensitive to small numbers, further improving classes that already have high recall brings little benefits to the harmonic mean, which makes sure no single class will be left behind. Also, now all classes are treated equally, forcing us to improve classes that have low recall no matter which subset it belongs to.

However, it is difficult to directly minimize the harmonic mean. We therefore propose a novel loss function that maximizes the geometric mean instead, which can be viewed as a surrogate. Our method serves as a simple plug-in that can be used together with both baseline and various state-of-the-art methods. We also propose an ensemble trick that uses the pre-trained and fine-tuned models together to make predictions during inference with nearly no extra cost. We are able to yield a more uniform distribution of recall across categories (i.e., no one left behind), which achieves higher harmonic mean of the recall while the (arithmetic) mean accuracy remains high, too.

In summary, our work has the following contributions:
\begin{itemize}
    \item We are the first to emphasize the importance of the correct recognition of all categories in long-tailed recognition.
    \item We propose a novel method that aims at increasing the harmonic mean of per-class recall as well as an ensemble trick that combines two existing models together during inference with nearly no extra cost.
    \item We experimented on three widely used benchmark datasets, which justify the effectiveness of our method in terms of both overall and worst per-class accuracy.
\end{itemize}

\section{Related Work}

Long-tailed recognition is a research field that aims at training models using imbalanced datasets. Since it is widely believed that the learning procedure is dominated by head classes, most existing works focus explicitly on improving the recognition of tail classes.

\subsection{Re-sampling and Re-weighting}

Re-sampling and re-weighting are two classical approaches. Re-sampling methods aim at balancing the training set by either over-sampling the tail classes~\cite{chawla2002smote,han2005borderline-smote,shen2016relay,park2022context-rich-minority} or under-sampling the head classes~\cite{he2009learning-from-imbalanced-data,drummond2003c4.5-class-imbalance}. Some methods also transfer statistics from the major classes to minor classes to obtain a balanced dataset~\cite{kim2020m2m,liu2019oltr}. One disadvantage of re-sampling methods is that it may lead to either over-fitting of tail classes or under-fitting of head classes. Re-weighting methods, on the other hand, give each instance a weight based on its true label when computing the loss~\cite{cao2019ldam,ren2020bsce,samuel2021distribution-robustness-loss}. The major drawback of re-weighting is that it makes the loss function hard to optimize.

\subsection{Two-Stage Decoupling}

Due to the shortcomings of re-sampling and re-weighting, various other methods are proposed recently. For example, two-stage methods like~\cite{kang2019decoupling,zhong2021improving-calibration,li2021ssd} propose to decouple the learning of features and classifiers and achieve impressive results. Our method is similar to them in the sense that we also re-train a classifier given an existing model. But our motivation is to design a simple and flexible method that can be applied to a wide variety of existing models, and we propose a novel loss function, while two-stage methods like cRT~\cite{kang2019decoupling} usually use the conventional cross entropy loss in the fine-tuning stage.

\subsection{Hybrid and Multiple Heads}

Hybrid methods, like~\cite{zhou2020bbn,wang2021ride}, make use of multiple heads to improve the recognition of different classes, which is similar to our proposed ensemble trick. But these methods require a joint training of multiple heads, together with a complex routing module that dynamically determines the head to use during inference, which increases the complexity of the model and makes it hard to use in practice. Our method, on the other hand, is simpler and requires no extra training nor complex modules to combine the predictions, which is different from them. 

\subsection{Miscellaneous}

There are many approaches that use knowledge distillation~\cite{he2021dive,li2021ssd} and contrastive learning~\cite{yang2020rethinking-the-value-of-labels,wang2021contrastive-learning-based-hybrid,cui2021paco,li2022targeted-supervised-contrastive,zhu2022balanced-contrastive} and achieve descent results. There are also methods that perform logit adjustment~\cite{zhang2021disalign,menon2021long-tail-learning-via-logit}, which formulate long-tailed recognition as a label distribution shift problem. Recently, there is also one work that tries to solve the problem by performing some regularization on the classifier's weight~\cite{alshammari2022weight-balancing}. They also visualize the recall of each class in their work and find the model is biased towards common classes, which is similar to our motivation. However, they do not emphasize the importance of correctly recognizing all classes, which is different from us.

\section{Method}

We will present our method in this section. First, we will introduce the notations used. Then we will describe the major framework of our method, which includes our proposed loss function and a simple ensemble trick.

\subsection{Notations}

The neural network can be represented as a non-linear function $\mathcal{F}$. Given the $i$-th training image $(\bm{x}^i,~y^i)$ from the imbalanced dataset that has $N$ images, where $\bm{x}^i \in \mathbb{R}^{H\times W\times D}$, a forward propagation through the network yields logit $\bm{o}^i=\mathcal{F}(\bm{x}^i)$. For a multi-class classification problem with $C$ categories, a softmax activation function is usually applied and the result can be represented as $\bm{\tilde{p}}^i=\mathrm{softmax}(\bm{o}^i)$, where $\bm{\tilde{p}}^i\in \mathbb{R}^{C}$.

\subsection{A Simple Plug-In Method}

Over the past few years, various methods~\cite{ren2020bsce,kang2019decoupling,cui2021paco} have been proposed to tackle the challenges in long tailed recognition. Although our evaluation results in ~\cref{tab:lowest-recall} reveal that no obvious improvements have been made regarding the worst-performing category, many of their design choices are critical and effective in long-tailed recognition. Therefore, in order to make full use of existing advances, we do not aim at building up a whole new training framework, but rather propose a simple plug-in method that can be applied to a wide range of existing methods.

Our method is simple. Given a model trained using either one of existing methods, we re-initialize the last FC layer and re-train it using our proposed loss function while the backbone is frozen. We borrow this idea from many two-stage decoupling methods~\cite{kang2019decoupling,zhong2021improving-calibration}, which is effective and flexible, making our method widely applicable. However, unlike normal decoupling methods that use the conventional cross entropy loss in the second stage, we propose a novel loss function to improve the worst-performing categories. Our proposed loss function, called GML (Geometric Mean Loss), is defined as
\begin{equation}
    \mathcal{L}_{\mathrm{GML}} = -\frac{1}{C}\sum_{c=1}^C\log \bm{\bar{p}}_c\,,
    \label{equation:loss-definition}
\end{equation}
where $\bm{\bar{p}}_c$ is computed as
\begin{equation}
    \bm{\bar{p}}_c=\frac{1}{N_{c}}\sum_{i=1}^{N_c}\bm{\tilde{p}}^i_{y^i}\,,
\end{equation}
which is the average of $\bm{\tilde{p}}^i_{y^i}$ across all training samples belonging to class $c$ in this mini-batch, and $N_c$ is the number of examples from this category in the mini-batch. Inspired by~\cite{ren2020bsce}, we perform re-weighting when we compute $\bm{\tilde{p}}^i$:
\begin{equation}
    \bm{\tilde{p}}^i_j = \frac{N_{j}\exp(o^i_j)}{\sum_{c=1}^CN_{c}\exp(o^i_c)}\,.
    \label{equation:reweighting}
\end{equation}

\subsection{Why GML Works?}

The definition of mean accuracy is
\begin{equation}
    \mathrm{Acc}=\frac{1}{N}\sum_{i=1}^N\vmathbb{1}(y^i=\arg\max \bm{\tilde{p}}^i)\,,
\end{equation}
where $\vmathbb{1}(\cdot)$ is the indicator function. On a balanced test set, it is the same as the arithmetic mean of per-class recall. We argue that maximizing the mean accuracy inevitably ignores some categories because arithmetic mean incurs little punishment to categories that have very small recall values. On the other hand, harmonic mean, defined as
\begin{equation}
    \mathrm{HM}(x_1,\dots,x_n)=\frac{n}{\frac{1}{x_1}+\dots+\frac{1}{x_n}}\,,
\end{equation}
incurs strong punishment to small values. Therefore, we believe it can be a better alternative to optimize for. However, harmonic mean is defined using reciprocal, making it hard and numeric unstable to be optimized. To this extent, we propose to maximize the geometric mean of per-class recall, defined as
\begin{equation}
    \mathrm{GM}(x_1,\dots,x_n)=\sqrt[n]{\left\vert x_1\times\dots\times x_n \right\vert}\,.
\end{equation}
As an illustration, $\frac{0.01+0.99}{2}=0.5$, $\mathrm{HM}(0.01,0.99)=0.02$, and $\mathrm{GM}(0.01,0.99)=0.10$. Both harmonic and geometric means are heavily affected by the small value (0.01), while the arithmetic mean is much less sensitive to it.

Specifically, given the per-class recall as $r_1,\dots,r_C$, we transform their geometric mean using a simple logarithm transformation:
\begin{align}
    \sqrt[C]{r_1\dots r_C}&=\exp\left(\log\left(r_1\dots r_C\right)^{1/C}\right)\\
    &=\exp\left(\frac{1}{C}\log\left(r_1\dots r_C\right)\right)\\
    &=\exp\left(\frac{1}{C}\sum_{i=1}^C\log r_i\right)\,.
\end{align}
Since recall is computed by computing the mean of a series of indicator functions and these indicator functions only take value $0$ or $1$ and is non-differentiable, we use $\bm{\bar{p}}_c$ as a surrogate for $r_c$ during training, resulting in
\begin{equation}
    \sqrt[C]{r_1\dots r_C}\propto \exp(-\mathcal{L}_{\mathrm{GML}})\,.
\end{equation}

So, by minimizing $\mathcal{L}_{\mathrm{GML}}$, we are effectively maximizing the geometric mean of per-class recall, thus being able to improve on the worst-performing categories.

\begin{figure}
    \centering
    \includegraphics[width=0.9\linewidth]{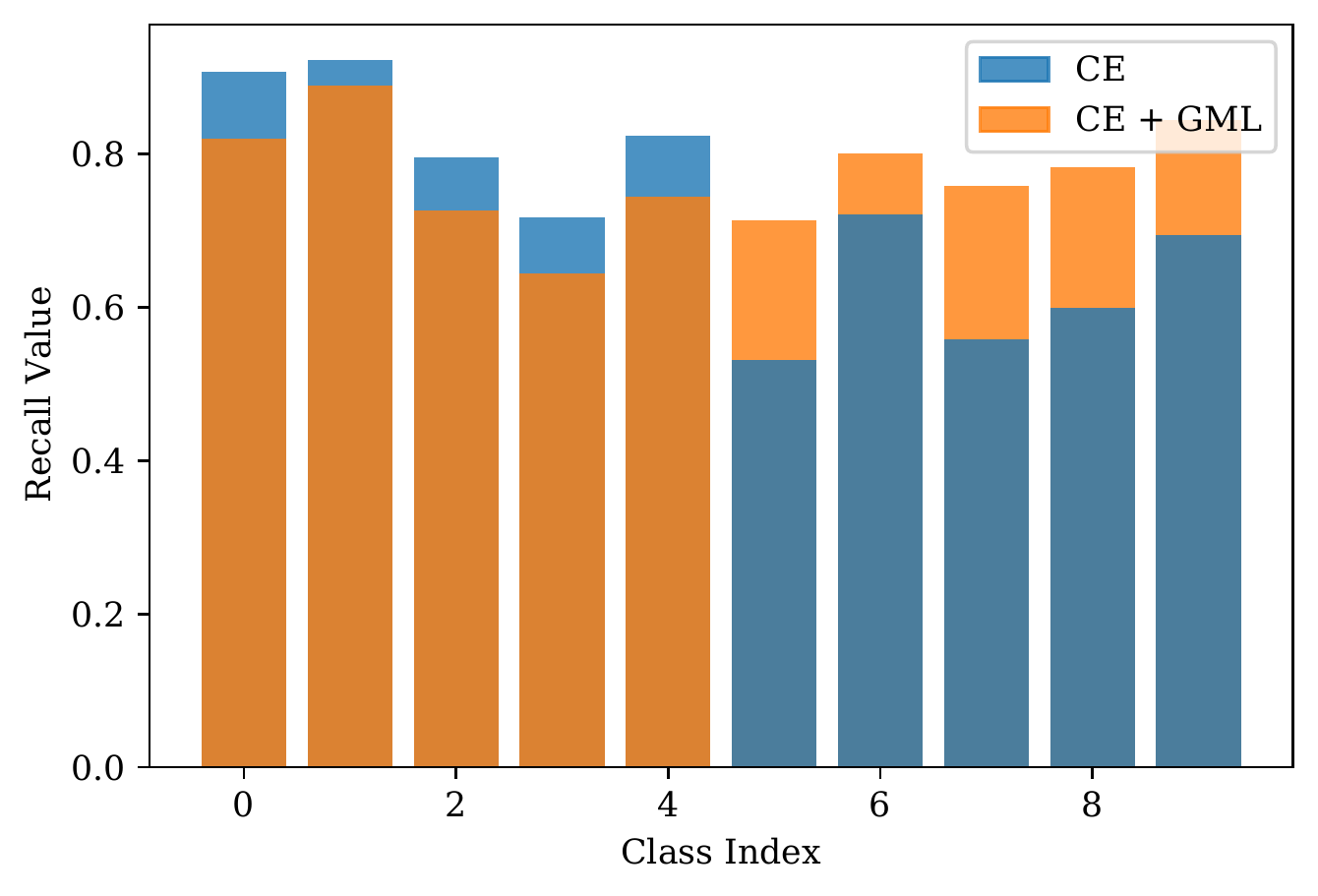}
    \caption{Bar plot of per-class recall on the imbalanced CIFAR10 (with imbalance ratio 100) before (`CE') and after (`CE+GML') the fine-tuning. For the fine-tuned model, the recall of the first 5 classes dropped while the recall of the latter 5 classes increased, motivating us to combine the prediction of both models.}
    \label{fig:recall-distribution-change}
\end{figure}

\subsection{Combining the Strength of Both Worlds}

By fine-tuning the pre-trained model with our proposed loss function, indeed we are able to obtain a more uniform distribution of recall values across categories. As is shown in~\cref{fig:recall-distribution-change}, the recall of classes that perform worse before are improved while the recall of classes that perform well before are dropped. Motivated by these dynamics, we then propose a simple ensemble trick that aims at combining the strength of both worlds.

Our idea is simple again. Since we re-initialize the classifier during fine-tuning and the backbone is kept frozen, the old classifier is still usable during inference. Let these two models be $\mathcal{F}_{\mathrm{old}}$ and $\mathcal{F}_{\mathrm{new}}$, respectively, we then make predictions by
\begin{equation}
    \bm{\tilde{p}}_{\mathrm{ensembled}}=\frac{\bm{\tilde{p}}_{\mathrm{new}} + \bm{\tilde{p}}_{\mathrm{old}}}{2}\,.
    \label{equation:ensemble-definition}
\end{equation}

In practice, since the calibration of some models can be poor~\cite{guo2017on-calibration,zhong2021improving-calibration} (e.g., the baseline using cross entropy), we need to manually calibrate the prediction before combining them. For simplicity, we choose temperature scaling to calibrate the model~\cite{guo2017on-calibration} and use two temperature variables $t_{\mathrm{new}}$ and $t_{\mathrm{old}}$ when computing $\bm{\tilde{p}}_{\mathrm{new}}$ and $\bm{\tilde{p}}_{\mathrm{old}}$ separately:
\begin{align}
    \bm{\tilde{p}}_{\mathrm{new}} &= \mathrm{softmax}\left(\frac{\bm{o}_{\mathrm{new}}}{t_{\mathrm{new}}}\right)\,,\label{equation:temperature-variables-new}\\
    \bm{\tilde{p}}_{\mathrm{old}} &= \mathrm{softmax}\left(\frac{\bm{o}_{\mathrm{old}}}{t_{\mathrm{old}}}\right)\,.
    \label{equation:temperature-variables-old}
\end{align}

One big advantage of our proposed ensemble trick is that it brings nearly no extra cost. Unlike many multi-head~\cite{zhou2020bbn} or multi-expert~\cite{wang2021ride} methods, we do not need extra training as we simply use the classifier that comes with the pre-trained model. During inference, the only extra cost is the fully connected layer to compute $\bm{\tilde{p}}_{\mathrm{old}}$, which is negligible. Also, no complex routing rules are needed.

The two temperature variables are the only two hyper-parameters introduced by our method, and they can be easily tuned as no training is required. Plus, as we will later show in the ablation studies, most values work perfectly fine so actually no painful tuning is needed. What's more, this ensemble trick is completely optional, our method works well enough even without it.

\subsection{Training Pipeline}

Our method is very similar to the two-stage decoupling methods~\cite{kang2019decoupling,zhong2021improving-calibration}. We design it in this way on purpose because we want it to be as flexible and simple as possible. During experiments, we found that our method is orthogonal to many design choices in state-of-the-art methods and is able to improve the worse-case performance when applied to various different pre-trained models. The overall training pipeline of our method consists of three stages:

\begin{enumerate}
    \item \textbf{Pre-training stage}: Obtain the pre-trained model from scratch using any one of the existing methods.
    \item \textbf{Fine-tuning stage}: Freeze the backbone and re-train the classifier using our proposed loss function.
    \item \textbf{An optional ensemble stage}: Calibrate the prediction of two classifiers and combine them additively.
\end{enumerate}

The overall training procedure is summarized in~\cref{algo:pipeline}.

\begin{algorithm}
    \renewcommand{\algorithmicrequire}{\textbf{Input:}}
	\caption{The overall training procedure}
	\begin{algorithmic}[1]
		\REQUIRE Training images $\bm{x}$ and their labels $y$.
		\STATE Randomly initialize the network and train it from scratch to obtain the pre-trained model.
		\STATE Use the pre-trained model to perform initialization. 
        \STATE Freeze the backbone and re-initialize the classifier.
		\STATE Fine-tune the model using the loss function defined as~\cref{equation:loss-definition} for a few epochs.
		\STATE During inference, combine the results from both classifiers as defined in~\cref{equation:ensemble-definition}.
	\end{algorithmic}
	\label{algo:pipeline}
\end{algorithm}

\section{Experiments}

\begin{table*}
    \centering
	\small
    \begin{tabular}{ccccc}
    \toprule
     Dataset & Number of Classes & \# Training Images & \# Test Images & Imbalance Ratio\\
     \midrule
     CIFAR100-LT~\cite{krizhevsky2009cifar}&100&10,847&10,000&100\\
     ImageNet-LT~\cite{deng2009imagenet,liu2019oltr}&1,000&115,846&50,000&256\\
     Places-LT~\cite{zhou2018places}&365&62,500&36,500&996\\
     \bottomrule
    \end{tabular}
    \tabcaption{Statistics of three imbalanced datasets used in our experiments.}
    \label{tab:dataset}
\end{table*}

In order to validate the effectiveness of our method, we have conducted experiments on three widely used benchmark datasets. In this section, we will present our experimental results. First, we will introduce the datasets used. Then we will describe various implementation details and evaluation metrics used. After that, we will compare our method with various other methods. Finally, we will present the results of some ablation studies.

\subsection{Datasets}

We use three widely used long-tailed image recognition datasets: CIFAR100-LT, ImageNet-LT and Places-LT. Some statistics about these three datasets are summarized in~\cref{tab:dataset}. Since CIFAR100~\cite{krizhevsky2009cifar}, ImageNet~\cite{deng2009imagenet} and Places~\cite{zhou2018places} are all balanced datasets, we follow previous works~\cite{zhou2020bbn,liu2019oltr} to down-sample their original training set. Details about the construction process can be found in the supplementary material.

\subsection{Implementation Details}

Following previous works~\cite{zhou2020bbn,liu2019oltr}, we use ResNet-32~\cite{he2016deep-residual}, ResNeXt-50~\cite{xie2017aggregated-residual} and ResNet-152 as the backbone network for CIFAR100-LT, ImageNet-LT and Places-LT, respectively. For the pre-training stage, if there exists released checkpoints, we use them directly. If not, we use the officially released codes and try to reproduce the results in our best effort without changing their original settings. This includes stuffs like hyper-parameters and data augmentations used. For the fine-tuning stage, detailed training settings can be found in the supplementary material.

\subsection{Evaluation Metrics and Comparison Methods}

\begin{table*}
    \centering
	\small
    \begin{tabular}{lcccc}
        \toprule
        Methods&\textcolor{gray}{Accuracy}&Geometric Mean&Harmonic Mean&Lowest Recall\\
        \midrule
        CE&\textcolor{gray}{38.74}&21.03*&2.05*&0.00\\
        BSCE~\cite{ren2020bsce}&\textcolor{gray}{42.24}&35.16&26.24&3.00\\
        \midrule
        LDAM~\cite{cao2019ldam}&\textcolor{gray}{43.51}&33.61&21.49&3.00\\
        LADE~\cite{hong2021lade}&\textcolor{gray}{44.39}&39.05&32.58&5.00\\
        DiVE~\cite{he2021dive}&\textcolor{gray}{45.11}&37.08&25.18&2.00\\
        MiSLAS~\cite{zhong2021improving-calibration}&\textcolor{gray}{47.05}&40.16&30.93&5.00\\
        RIDE~\cite{wang2021ride}&\textcolor{gray}{48.64}&38.71&23.86&2.00\\
        PaCo~\cite{cui2021paco}&\textcolor{gray}{51.24}&45.29&36.42&5.00\\
        \midrule
        CE + GML&\textcolor{gray}{41.06}&36.59&31.26&6.00\\
        PaCo + GML&\textcolor{gray}{50.53}&45.47&39.20&9.00\\
        PaCo + GML (Ensemble)&\textcolor{gray}{49.82}&\textbf{45.70}&\textbf{41.02}&\textbf{15.00}\\
        \bottomrule
    \end{tabular}
    \caption{Results on the CIFAR100-LT dataset with imbalance ratio 100. Numbers with * are computed by substituting zero elements with a small number ($10^{-3}$) or else the geometric and harmonic mean will all be zero.}
    \label{tab:CIFAR100LT-res}
\end{table*}

\begin{table*}
    \centering
	\small
    \begin{tabular}{lcccc}
        \toprule
        Methods&\textcolor{gray}{Accuracy}&Geometric Mean&Harmonic Mean&Lowest Recall\\
        \midrule
        CE&\textcolor{gray}{43.90}&23.25*&1.25*&0.00\\
        BSCE~\cite{ren2020bsce}&\textcolor{gray}{50.48}&42.32*&13.74*&0.00\\
        \midrule
        cRT~\cite{kang2019decoupling}&\textcolor{gray}{49.64}&41.35*&13.82*&0.00\\
        DiVE~\cite{he2021dive}&\textcolor{gray}{53.63}&45.49*&12.76*&0.00\\
        RIDE~\cite{wang2021ride}&\textcolor{gray}{55.69}&47.56*&17.32*&0.00\\
        PaCo~\cite{cui2021paco}&\textcolor{gray}{58.53}&51.32*&21.79*&0.00\\
        \midrule
        CE + GML&\textcolor{gray}{45.61}&39.82&31.67&2.00\\
        PaCo + GML&\textcolor{gray}{55.57}&50.67&43.71&2.00\\
        PaCo + GML (Ensemble)&\textcolor{gray}{57.22}&\textbf{52.29}&\textbf{44.80}&\textbf{2.00}\\
        \bottomrule
    \end{tabular}
    \caption{Results on the ImageNet-LT dataset. Numbers with * are computed by substituting zero elements with a small number ($10^{-3}$) or else the geometric and harmonic mean will all be zero.}
    \label{tab:ImageNet-LT-res}
\end{table*}

\begin{table*}
    \centering
	\small
    \begin{tabular}{lcccc}
        \toprule
        Methods&\textcolor{gray}{Accuracy}&Geometric Mean&Harmonic Mean&Lowest Recall\\
        \midrule
        CE&\textcolor{gray}{28.71}&12.11*&0.73*&0.00\\
        BSCE~\cite{ren2020bsce}&\textcolor{gray}{37.18}&29.30*&5.64*&0.00\\
        \midrule
        PaCo~\cite{cui2021paco}&\textcolor{gray}{40.45}&27.88*&2.53*&0.00\\
        MiSLAS~\cite{zhong2021improving-calibration}&\textcolor{gray}{40.48}&35.53&28.97&3.00\\
        \midrule
        CE + GML&\textcolor{gray}{36.82}&29.30*&8.00*&0.00\\
        MiSLAS + GML&\textcolor{gray}{39.90}&35.41&29.98&5.00\\
        MiSLAS + GML (Ensemble)&\textcolor{gray}{40.37}&\textbf{35.92}&\textbf{30.51}&\textbf{5.00}\\
        \bottomrule
    \end{tabular}
    \caption{Results on the Places-LT dataset. Numbers with * are computed by substituting zero elements with a small number ($10^{-3}$) or else the geometric and harmonic mean will all be zero. Note that to plug GML into MiSLAS, since MiSLAS is also a two-stage method, we think it's fairer to use GML together with their proposed label aware smoothing loss.}
    \label{tab:Places-LT-res}
\end{table*}

The convention in long-tailed recognition research is to split the classes into three subsets based on the number of training images~\cite{kang2019decoupling,liu2019oltr}. Typically, ``Head'' denotes classes that have more than 100 images, ``Few'' denotes classes that have less than 20 images and all other classes are denoted as ``Medium''. The accuracy within each subset is often reported along with the overall accuracy to justify the claim that the performance improvement mainly comes from tail classes. And according to our observation in~\cref{sec:introduction}, we argue that such an evaluation scheme can be problematic and we propose to first compute the recall within each category and then compute their harmonic mean. In our experiments, since our method optimizes geometric mean of per-class recall as a surrogate, we also report the geometric mean. Besides, we report the overall mean accuracy as well.

We compare our methods with both baseline methods like simple cross entropy loss and various state-of-the-arts methods~\cite{ren2020bsce,he2021dive,wang2021ride,cui2021paco}. Many of these methods have very different training settings and strictly speaking, they are not directly comparable. But since our method is a simple plug-in method, the main focus is to quantify the improvements brought by our method compared to its respective baseline result.

\subsection{Main Comparisons}

We validate our method on three benchmark datasets and present the results on each dataset here separately.

\subsubsection{CIFAR100-LT}

\cref{tab:CIFAR100LT-res} shows the experimental results on the CIFAR100-LT dataset. CIFAR100-LT has three different versions with different imbalance ratios and here we only consider the one with imbalance ratio 100. As we can see from the table, although various different methods have been proposed in recent years and there are rapid improvements on the overall accuracy, the performance of the worst-performing category is still relatively poor. The geometric and harmonic mean of per-class recall also justify our claim. We apply our method to both the baseline method CE and state-of-the-art method PaCo~\cite{cui2021paco} on this dataset. When our method is applied to CE and PaCo, we are able to improve the lowest recall from 0.00 and 5.00 to 6.00 and 9.00 respectively. And if we further apply our proposed ensemble trick, we are able to improve the lowest recall to 15.00, achieving state-of-the-art performance in both geometric mean and harmonic mean while keeping the overall accuracy roughly unchanged. 

It is worth noting that even the simplest CE+GML outperforms all existing methods in terms of lowest recall, that is, in terms of ``no one left behind''.

\subsubsection{ImageNet-LT and Places-LT}

\cref{tab:ImageNet-LT-res} shows the result on ImageNet-LT. The overall observation is similar to that on CIFAR100-LT, except that since this dataset is larger and harder, \emph{the lowest recall of all current methods are zero}, making the harmonic mean of per-class recall very low, even when a small number substitutes zero per-class recalls. Our method, on the other hand, greatly improves the harmonic mean by successfully improving the lowest recall value. All categories have per-class recall higher than 0 when our method is used, even in the simplest CE+GML.

\cref{tab:Places-LT-res} shows the result on Places-LT. Current state-of-the-art on this dataset is MiSLAS~\cite{zhong2021improving-calibration}, so we apply our method upon it and the simple baseline CE. Again, our method successfully improves the lowest recall, as well as the harmonic mean of per-class recall. And when the ensemble trick is further applied, we are able to achieve new state-of-the-art results.

\subsubsection{Universality}

Since we want our method to be a universal plug-in, in this subsection we present the results of the experiment of adding our method on top of various different methods. The experiment is conducted on CIFAR100-LT, and the result is shown in~\cref{tab:universality}. Note that here we do not perform any ensemble. As we can see, our method is applicable to various different methods, and is able to achieve consistent improvements. However, we do observed that there exists some methods, like DiVE~\cite{he2021dive}, where we failed to achieve obvious improvements. More details about this can be found when we discuss limitations in~\cref{sec:limitations}.

\begin{table}
    \centering
    \resizebox{\columnwidth}{!}{
    \begin{tabular}{lccc}
        \toprule
        Methods&G-Mean&H-Mean&Lowest Recall\\
        \midrule
        CE&21.03*&2.05*&0.00\\
        CE + GML&36.59~(\textcolor{blue}{15.56$\uparrow$})&31.26~(\textcolor{blue}{29.21$\uparrow$})&6.00~(\textcolor{blue}{6.00$\uparrow$})\\
        \midrule
        BSCE&35.16&26.24&3.00\\
        BSCE + GML&36.52~(\textcolor{blue}{1.36$\uparrow$})&30.88~(\textcolor{blue}{4.64$\uparrow$})&7.00~(\textcolor{blue}{4.00$\uparrow$})\\
        \midrule
        MiSLAS&40.16&30.93&5.00\\
        MiSLAS + GML&40.90~(\textcolor{blue}{0.74$\uparrow$})&36.49~(\textcolor{blue}{5.56$\uparrow$})&11.00~(\textcolor{blue}{6.00$\uparrow$})\\
        \midrule
        PaCo&45.29&36.42&5.00\\
        PaCo + GML&45.47~(\textcolor{blue}{0.18$\uparrow$})&39.20~(\textcolor{blue}{2.78$\uparrow$})&9.00~(\textcolor{blue}{4.00$\uparrow$})\\
        \bottomrule
    \end{tabular}
    }
    \caption{Results on the CIFAR100-LT dataset with imbalance ratio 100. Our proposed method is applicable to various methods. ``G-Mean'' is short for ``Geometric Mean'' and ``H-Mean'' is short for ``Harmonic Mean''.}
    \label{tab:universality}
\end{table}

\subsection{Ablation Studies}

\begin{table}
    \centering
    \resizebox{\columnwidth}{!}{
    \begin{tabular}{ccccc}
        \toprule
        $t_{\mathrm{old}}$&$t_{\mathrm{new}}$&Geometric Mean&Harmonic Mean&Lowest Recall\\
        \midrule
        1&1&45.70&\textbf{41.02}&\textbf{15.00}\\
        1&2&45.58&38.86&8.00\\
        1&3&44.68&36.09&5.00\\
        2&1&44.39&40.22&\textbf{15.00}\\
        2&2&45.61&41.01&\textbf{15.00}\\
        2&3&\textbf{46.06}&40.40&11.00\\
        3&1&43.73&39.63&\textbf{15.00}\\
        \bottomrule
    \end{tabular}
    }
    \caption{Ablation on the effects of temperature variables when we apply our ensemble method upon PaCo~\cite{cui2021paco}.}
    \label{tab:effects-temperature-variables}
\end{table}

For ablation studies, if not otherwise specified, all experiments are conducted on CIFAR100-LT (with imbalance ratio 100) under the default training setting.

\subsubsection{Effects of Two Temperature Variables in the Ensemble Stage}

When we design our method, we'd like to keep the number of new hyper-parameters as less as possible in order for it to be readily applicable in real-world applications. The only two hyper-parameters introduced by our method are in the optional ensemble stage. Here we study the effects of choosing different values of them. Specifically, we change the value of the temperature variables shown in~\cref{equation:temperature-variables-new} and~\cref{equation:temperature-variables-old} in the ensemble stage when we apply our method to PaCo~\cite{cui2021paco}. The results are shown in~\cref{tab:effects-temperature-variables}. In our experiments on CIFAR100-LT, we set $t_{\mathrm{old}}=1$ and $t_{\mathrm{new}}=1$. And as we can see from the table, the value of $t_{\mathrm{old}}$ does not affect the result much while $t_{\mathrm{new}}$ is somewhat important, especially for improving the lowest recall---higher $t_{\mathrm{new}}$ will hurt it. This is natural because higher $t$ results in a smoother distribution and less information is preserved. Since the newly trained classifier is crucial for improving the lowest recall, we need to preserve more information from it. 

Generally speaking, since these two hyper-parameters are introduced for calibration, for different methods, we may need to tune them respectively because the calibrations of different methods are known to vary a lot in long-tailed recognition problems~\cite{zhong2021improving-calibration}. But usually this will not cause a headache, because the ensemble stage does not involve any training and thus can be tuned very quickly.

\subsubsection{Re-weighting in GML}

\begin{table}
    \centering
    \resizebox{\columnwidth}{!}{
    \begin{tabular}{lccc}
        \toprule
        Methods&G-Mean&H-Mean&Lowest Recall\\
        \midrule
        CE + GML&\textbf{36.59}&\textbf{31.26}&\textbf{6.00}\\
        \midrule
        (w/o re-weighting)&25.45&5.11&0.00\\
        (w/o re-weighting, re-sampling)&32.32&23.05&4.00\\
        \bottomrule
    \end{tabular}
    }
    \caption{Ablation on the re-weighting design in GML.}
    \label{tab:ablate-reweight}
\end{table}

As shown in~\cref{equation:reweighting}, re-weighting is applied in our proposed GML, here we ablate this design. The results are shown in~\cref{tab:ablate-reweight}. As we can see, the performance will be relatively poor without re-weighting. And just like many two-stage decoupling methods~\cite{kang2019decoupling}, re-sampling during fine-tuning can help, but is inferior to re-weighting.

\subsubsection{Training from Scratch using GML}

\begin{table}
    \centering
    \resizebox{\columnwidth}{!}{
    \begin{tabular}{lccc}
        \toprule
        Methods&G-Mean&H-Mean&Lowest Recall\\
        \midrule
        CE + GML&36.59&31.26&6.00\\
        PaCo + GML&\textbf{45.47}&\textbf{39.20}&\textbf{9.00}\\
        \midrule
        GML (train from scratch)&36.63&30.86&\textbf{9.00}\\
        \bottomrule
    \end{tabular}
    }
    \caption{Training from scratch using GML.}
    \label{tab:train-from-scratch}
\end{table}

Since our main purpose is to propose a simple plug-in method that is widely applicable to various existing methods, we use GML only to re-train the classifier of an existing model. However, GML can also be used to train a model from scratch. The results are shown in~\cref{tab:train-from-scratch}. As we can see, the result is comparable to the result when GML is attached to CE, but inferior to the result when GML is applied to SOTA methods like PaCo, just as expected.

\subsubsection{Visualization of the Per-Class Recall}

\begin{figure}
    \centering
    \includegraphics[width=0.9\linewidth]{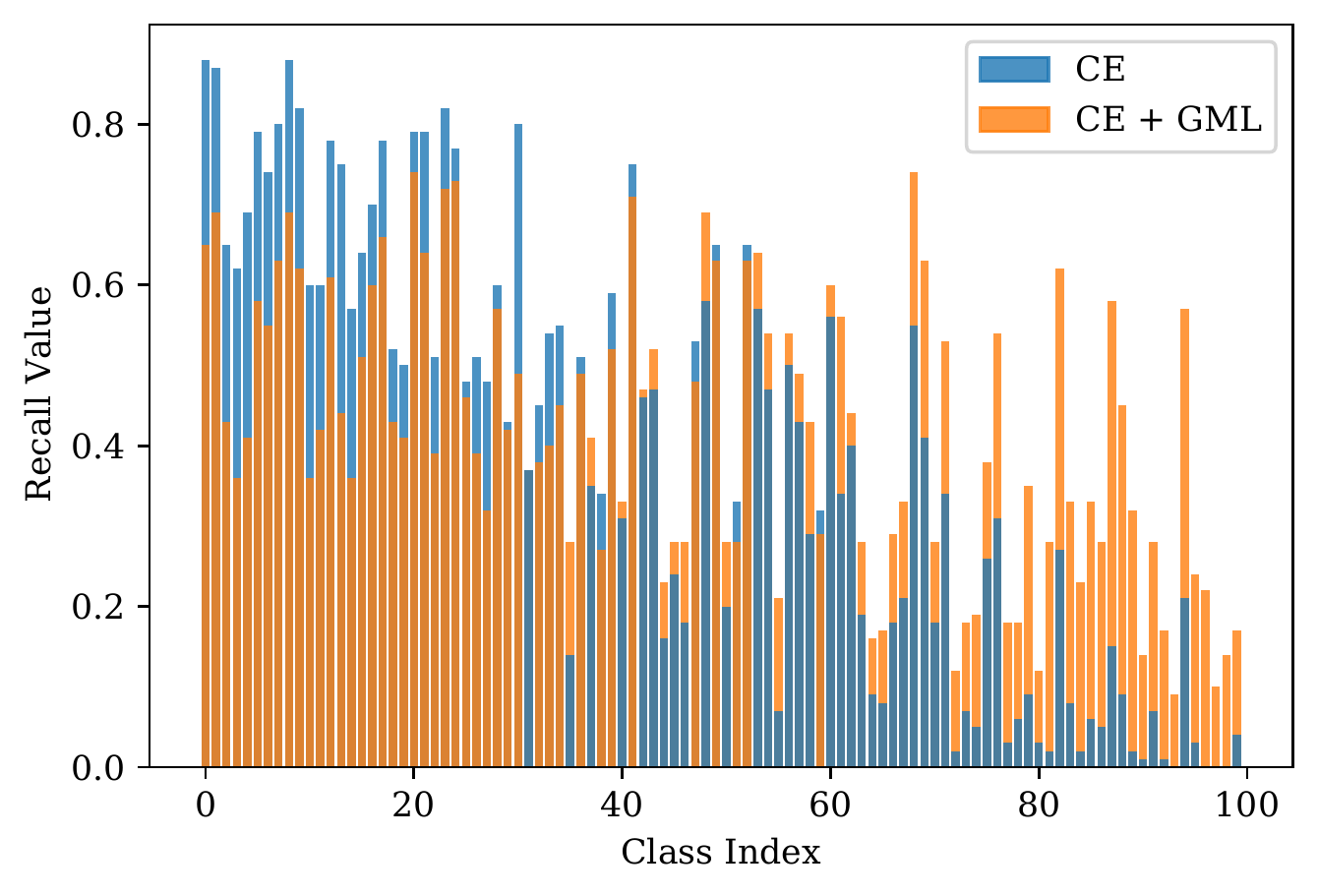}
    \caption{Bar plot of per-class recall on the imbalanced CIFAR100 (with imbalance ratio 100) before and after the fine-tuning. No ensemble is performed. Our method focuses on improving the recognition of classes that have low recall values, making sure no category is left behind.}
    \label{fig:recall-distribution-change-CIFAR100LT}
\end{figure}

In~\cref{fig:per-class-recall}, we visualize the per-class recall of model trained on the imbalanced dataset, and find they vary a lot from class to class. Since our major motivation is to make sure all classes are equally treated so that no category will be left behind, here we visualize the change of per-class recall after applying our method. We perform visualizing using two datasets, CIFAR10-LT and CIFAR100-LT, both have imbalance ratio 100. The visualization results on CIFAR10-LT and CIFAR100-LT are shown in~\cref{fig:recall-distribution-change} and~\cref{fig:recall-distribution-change-CIFAR100LT}, respectively. Note that here we do not perform the ensemble. As we can clearly see in both figures, our fine-tuned model has a more uniform distribution of per-class recall. These two figures justify the effectiveness of our proposed method.

\section{Limitations and Broader  Impacts}\label{sec:limitations}

Although our method is applicable to a wide range of existing methods, we do find that when applied to some methods, the improvement is less obvious compared to others. For example, our framework does not cope well with methods like BSCE~\cite{ren2020bsce} and DiVE~\cite{he2021dive}. Interestingly, although DiVE is based on knowledge distillation, BSCE is used in DiVE as well to train both the teacher and student model so we attribute the failure to the fact that these method all re-weight the loss function. Two-stage decoupling methods like cRT~\cite{kang2019decoupling} are also known to not perform well when the first stage training involves re-weighting or re-sampling. The author argues it is because re-sampling and re-weighting have negative effects on feature learning and we believe that argument applies to our method as well. Another thing worth noting here is that we do not report the experimental results on iNaturalist2018~\cite{cui2018inat} because we find it generally not suitable in our settings. iNaturalist2018 is another widely used benchmark dataset in long-tailed recognition. It is large-scale, consisting of 437.5K training images from 8142 classes. But for the test split, it only has 24K images, so each class only has 3 images. Since our major motivation is to make sure no category is left behind, it is generally not meaningful to look at the per-class recall in iNaturalist2018 because the variance of the result can be very large due to the very small number of test images. However, in real-world applications, it is generally rare to have such small number of test images.

For broader impacts, we believe our research have positive impacts on reducing the bias of model training on an imbalanced dataset. Machine learning models are known to be biased if there is lack of training images from some categories. We believe our work will be helpful to achieve better fairness by making sure all categories are equally treated.

\section{Conclusions}

In this paper, we focused on improving some of the worst-performing classes in long-tailed recognition. We found that when trained with an imbalanced dataset, the per-class recall of the model varies a lot from class to class. Previous convention reports the arithmetic mean of per-class recall, but we argued such an evaluation scheme can be problematic. First, arithmetic mean only incurs little penalty to small numbers, making it too coarse to reflect whether there are categories that are left behind (i.e., with close to zero per-class accuracy). Second, it is not necessarily true that classes in the ``Few'' category perform worse than those in ``Many'' or ``Medium'', so focusing on improving ``Few'' accuracy alone is not enough. For these reasons, we argued that we should pay more attention to the harmonic mean of per-class recall and we therefore proposed a novel method that aims at improving the lowest recall and the harmonic mean of recalls. Our method is a simple plug-in that is applicable to a wide range of existing methods. Specifically, our method consists of three stages, in the first stage, we use any existing method to train a model from scratch. Then we re-train the classifier using our proposed novel GML loss function. Finally, we propose a simple ensemble trick that can be used to combine the predictions from the two classifiers with nearly no extra cost. We validated the effectiveness of our method on three widely used benchmark datasets, and witnessed consistent improvements on the harmonic mean of recalls and lowest recall value, while the overall accuracy still remains high. By visualizing the distribution of per-class recall values of the fine-tuned model, we found our model indeed achieved a more balanced distribution.

{\small
\bibliographystyle{ieee_fullname}
\bibliography{egbib}

\begin{thebibliography}{10}\itemsep=-1pt

\bibitem{alshammari2022weight-balancing}
Shaden Alshammari, Yu-Xiong Wang, Deva Ramanan, and Shu Kong.
\newblock Long-tailed recognition via weight balancing.
\newblock In {\em Proceedings of the IEEE/CVF Conference on Computer Vision and
  Pattern Recognition}, pages 6897--6907, June 2022.

\bibitem{cao2019ldam}
Kaidi Cao, Colin Wei, Adrien Gaidon, Nikos Arechiga, and Tengyu Ma.
\newblock Learning imbalanced datasets with label-distribution-aware margin
  loss.
\newblock In {\em Advances in Neural Information Processing Systems 32}, pages
  1567--1578, December 2019.

\bibitem{chawla2002smote}
Nitesh~V Chawla, Kevin~W Bowyer, Lawrence~O Hall, and W~Philip Kegelmeyer.
\newblock {SMOTE}: Synthetic minority over-sampling technique.
\newblock {\em Journal of Artificial Intelligence Research}, 16:321--357, 2002.

\bibitem{cui2021paco}
Jiequan Cui, Zhisheng Zhong, Shu Liu, Bei Yu, and Jiaya Jia.
\newblock Parametric contrastive learning.
\newblock In {\em Proceedings of the IEEE/CVF International Conference on
  Computer Vision}, pages 715--724, October 2021.

\bibitem{cui2018inat}
Yin Cui, Yang Song, Chen Sun, Andrew Howard, and Serge Belongie.
\newblock Large scale fine-grained categorization and domain-specific transfer
  learning.
\newblock In {\em Proceedings of the IEEE Conference on Computer Vision and
  Pattern Recognition}, pages 4109--4118, June 2018.

\bibitem{deng2009imagenet}
Jia Deng, Wei Dong, Richard Socher, Li-Jia Li, Kai Li, and Li Fei-Fei.
\newblock {ImageNet}: A large-scale hierarchical image database.
\newblock In {\em 2009 IEEE Conference on Computer Vision and Pattern
  Recognition}, pages 248--255, June 2009.

\bibitem{drummond2003c4.5-class-imbalance}
Chris Drummond and Robert~C. Holte.
\newblock C4.5, class imbalance, and cost sensitivity: Why under-sampling beats
  over-sampling.
\newblock In {\em Workshop on Learning from Imbalanced Data Sets II}, 2003.

\bibitem{guo2017on-calibration}
Chuan Guo, Geoff Pleiss, Yu Sun, and Kilian~Q. Weinberger.
\newblock On calibration of modern neural networks.
\newblock In {\em Proceedings of the 34th International Conference on Machine
  Learning}, pages 1321--1330, August 2017.

\bibitem{han2005borderline-smote}
Hui Han, Wen-Yuan Wang, and Bing-Huan Mao.
\newblock {Borderline-SMOTE}: A new over-sampling method in imbalanced data
  sets learning.
\newblock In {\em Advances in Intelligent Computing}, pages 878--887, August
  2005.

\bibitem{he2009learning-from-imbalanced-data}
Haibo He and Edwardo~A. Garcia.
\newblock Learning from imbalanced data.
\newblock {\em IEEE Transactions on Knowledge and Data Engineering},
  21(9):1263--1284, 2009.

\bibitem{he2016deep-residual}
Kaiming He, Xiangyu Zhang, Shaoqing Ren, and Jian Sun.
\newblock Deep residual learning for image recognition.
\newblock In {\em Proceedings of the IEEE Conference on Computer Vision and
  Pattern Recognition}, pages 770--778, June 2016.

\bibitem{he2021dive}
Yin-Yin He, Jianxin Wu, and Xiu-Shen Wei.
\newblock Distilling virtual examples for long-tailed recognition.
\newblock In {\em Proceedings of the IEEE/CVF International Conference on
  Computer Vision}, pages 235--244, October 2021.

\bibitem{hong2021lade}
Youngkyu Hong, Seungju Han, Kwanghee Choi, Seokjun Seo, Beomsu Kim, and Buru
  Chang.
\newblock Disentangling label distribution for long-tailed visual recognition.
\newblock In {\em Proceedings of the IEEE/CVF Conference on Computer Vision and
  Pattern Recognition}, pages 6626--6636, June 2021.

\bibitem{kang2019decoupling}
Bingyi Kang, Saining Xie, Marcus Rohrbach, Zhicheng Yan, Albert Gordo, Jiashi
  Feng, and Yannis Kalantidis.
\newblock Decoupling representation and classifier for long-tailed recognition.
\newblock In {\em International Conference on Learning Representations}, April
  2020.

\bibitem{kim2020m2m}
Jaehyung Kim, Jongheon Jeong, and Jinwoo Shin.
\newblock M2m: Imbalanced classification via major-to-minor translation.
\newblock In {\em Proceedings of the IEEE/CVF Conference on Computer Vision and
  Pattern Recognition}, pages 13896--13905, June 2020.

\bibitem{krizhevsky2009cifar}
Alex Krizhevsky.
\newblock Learning multiple layers of features from tiny images.
\newblock Technical report, University of Toronto, 2009.
\newblock
  \nolinkurl{https://www.cs.toronto.edu/~kriz/learning-features-2009-TR.pdf}.

\bibitem{li2022targeted-supervised-contrastive}
Tianhong Li, Peng Cao, Yuan Yuan, Lijie Fan, Yuzhe Yang, Rogerio~S. Feris,
  Piotr Indyk, and Dina Katabi.
\newblock Targeted supervised contrastive learning for long-tailed recognition.
\newblock In {\em Proceedings of the IEEE/CVF Conference on Computer Vision and
  Pattern Recognition}, pages 6918--6928, June 2022.

\bibitem{li2021ssd}
Tianhao Li, Limin Wang, and Gangshan Wu.
\newblock Self supervision to distillation for long-tailed visual recognition.
\newblock In {\em Proceedings of the IEEE/CVF International Conference on
  Computer Vision}, pages 630--639, October 2021.

\bibitem{liu2019oltr}
Ziwei Liu, Zhongqi Miao, Xiaohang Zhan, Jiayun Wang, Boqing Gong, and Stella~X.
  Yu.
\newblock Large-scale long-tailed recognition in an open world.
\newblock In {\em Proceedings of the IEEE/CVF Conference on Computer Vision and
  Pattern Recognition}, pages 2537--2546, June 2019.

\bibitem{menon2021long-tail-learning-via-logit}
Aditya~Krishna Menon, Sadeep Jayasumana, Ankit~Singh Rawat, Himanshu Jain,
  Andreas Veit, and Sanjiv Kumar.
\newblock Long-tail learning via logit adjustment.
\newblock In {\em International Conference on Learning Representations}, May
  2021.

\bibitem{park2022context-rich-minority}
Seulki Park, Youngkyu Hong, Byeongho Heo, Sangdoo Yun, and Jin~Young Choi.
\newblock The majority can help the minority: Context-rich minority
  oversampling for long-tailed classification.
\newblock In {\em Proceedings of the IEEE/CVF Conference on Computer Vision and
  Pattern Recognition}, pages 6887--6896, June 2022.

\bibitem{ren2020bsce}
Jiawei Ren, Cunjun Yu, Shunan Sheng, Xiao Ma, Haiyu Zhao, Shuai Yi, and
  Hongsheng Li.
\newblock Balanced meta-softmax for long-tailed visual recognition.
\newblock In {\em Advances in Neural Information Processing Systems 33}, pages
  4175--4186, December 2020.

\bibitem{samuel2021distribution-robustness-loss}
Dvir Samuel and Gal Chechik.
\newblock Distributional robustness loss for long-tail learning.
\newblock In {\em Proceedings of the IEEE/CVF International Conference on
  Computer Vision}, pages 9495--9504, October 2021.

\bibitem{shen2016relay}
Li Shen, Zhouchen Lin, and Qingming Huang.
\newblock Relay backpropagation for effective learning of deep convolutional
  neural networks.
\newblock In {\em Proceedings of the European Conference on Computer Vision},
  pages 467--482, October 2016.

\bibitem{wang2021contrastive-learning-based-hybrid}
Peng Wang, Kai Han, Xiu-Shen Wei, Lei Zhang, and Lei Wang.
\newblock Contrastive learning based hybrid networks for long-tailed image
  classification.
\newblock In {\em Proceedings of the IEEE/CVF Conference on Computer Vision and
  Pattern Recognition}, pages 943--952, June 2021.

\bibitem{wang2021ride}
Xudong Wang, Long Lian, Zhongqi Miao, Ziwei Liu, and Stella Yu.
\newblock Long-tailed recognition by routing diverse distribution-aware
  experts.
\newblock In {\em International Conference on Learning Representations}, May
  2021.

\bibitem{xie2017aggregated-residual}
Saining Xie, Ross Girshick, Piotr Dollar, Zhuowen Tu, and Kaiming He.
\newblock Aggregated residual transformations for deep neural networks.
\newblock In {\em Proceedings of the IEEE Conference on Computer Vision and
  Pattern Recognition}, pages 1492--1500, July 2017.

\bibitem{yang2020rethinking-the-value-of-labels}
Yuzhe Yang and Zhi Xu.
\newblock Rethinking the value of labels for improving class-imbalanced
  learning.
\newblock In {\em Advances in Neural Information Processing Systems 33}, pages
  19290--19301, December 2020.

\bibitem{zhang2021disalign}
Songyang Zhang, Zeming Li, Shipeng Yan, Xuming He, and Jian Sun.
\newblock Distribution alignment: A unified framework for long-tail visual
  recognition.
\newblock In {\em Proceedings of the IEEE/CVF Conference on Computer Vision and
  Pattern Recognition}, pages 2361--2370, June 2021.

\bibitem{zhong2021improving-calibration}
Zhisheng Zhong, Jiequan Cui, Shu Liu, and Jiaya Jia.
\newblock Improving calibration for long-tailed recognition.
\newblock In {\em Proceedings of the IEEE/CVF Conference on Computer Vision and
  Pattern Recognition}, pages 16489--16498, June 2021.

\bibitem{zhou2020bbn}
Boyan Zhou, Quan Cui, Xiu-Shen Wei, and Zhao-Min Chen.
\newblock {BBN}: Bilateral-branch network with cumulative learning for
  long-tailed visual recognition.
\newblock In {\em Proceedings of the IEEE/CVF Conference on Computer Vision and
  Pattern Recognition}, pages 9719--9728, June 2020.

\bibitem{zhou2018places}
Bolei Zhou, Agata Lapedriza, Aditya Khosla, Aude Oliva, and Antonio Torralba.
\newblock Places: A 10 million image database for scene recognition.
\newblock {\em IEEE Transactions on Pattern Analysis and Machine Intelligence},
  40(6):1452--1464, 2018.

\bibitem{zhu2022balanced-contrastive}
Jianggang Zhu, Zheng Wang, Jingjing Chen, Yi-Ping~Phoebe Chen, and Yu-Gang
  Jiang.
\newblock Balanced contrastive learning for long-tailed visual recognition.
\newblock In {\em Proceedings of the IEEE/CVF Conference on Computer Vision and
  Pattern Recognition}, pages 6908--6917, June 2022.

\end{thebibliography}
}

\clearpage
\appendix
\section{More Details about the Datasets Used}

The imbalance ratio of a dataset is defined as the number of training images of the most frequent class divide by the number of images of the least frequent class.

\textbf{CIFAR100-LT.} CIFAR100~\cite{krizhevsky2009cifar} is a balanced dataset for image recognition, which has 50,000 training images and 10,000 test images from 100 categories. The CIFAR100-LT dataset used in our experiments are obtained by down-sampling the original training set while keeping the test set unchanged. Following Zhou~\etal~\cite{zhou2020bbn}, we use the exponential function $N_i=N_0\times \mu^i$ to determine the number of training images for each category, where $N_0=500$. By varying $\mu^i$, we are able to construct datasets with different imbalance ratios. In our experiments, we only use the one with imbalance ratio 100.

\textbf{ImageNet-LT and Places-LT.} ImageNet~\cite{deng2009imagenet} and Places~\cite{zhou2018places} are also two balanced dataset. Unlike CIFAR, these two datasets have larger scale and are more difficult. ImageNet-LT and Places-LT are their long-tailed version constructed by Liu~\etal~\cite{liu2019oltr}. The number of training images for each class is determined using the Pareto distribution with a power value $\alpha=6$. Their original test sets are left unchanged.

\section{Implementation Details of the Fine-tuning Stage}

\textbf{CIFAR100-LT.} For data augmentation, we randomly crop a $32\times 32$ patch from the original image or its horizontal flip with 4 pixels padded on each side. We use the stochastic gradient descent (SGD) to optimize the network with momentum of $0.9$ and weight decay of $5\times 10^{-4}$. We train the model for $40$ epochs. The initial learning rate is set to $5\times 10^{-2}$ and decrease it at the $10^{\mathrm{th}}$ epoch by $0.2$. We use a batch size of $128$.

\textbf{ImageNet-LT and Places-LT}. For data augmentation, we resize the image by setting the shorter side to $256$ and then take a random crop of $224\times 224$ from it or its horizontal flip. Finally, color jittering is applied. We train our model for $40$ epochs with a batch size of $512$. We use stochastic gradient descent (SGD) with momentum of $0.9$ and weight decay of $5\times 10^{-4}$. The initial learning rate is set to $5\times 10^{-2}$ and is decreased at the $20^{\mathrm{th}}$ epoch by $0.2$. For Places-LT, when applied to MiSLAS~\cite{zhong2021improving-calibration}, since MiSLAS is a two-stage method, we find it fairer to also apply their proposed label aware smoothing loss in the fine-tuning stage. So when computing the loss function, we combine two loss functions together as $\mathcal{L} = \lambda * \mathcal{L}_{\mathrm{GML}} + \mathcal{L}_{\mathrm{LAS}}$. During the experiment, we simply use $\lambda=1$ without tuning it.

\section{Additional Ablation Studies}

We present some additional ablation studies here.

\subsection{Better Baselines}

Since our method requires re-training the classifier, the model is essentially trained longer. To better understand the performance improvement, here we conduct some experiments on CIFAR100-LT that serve as better baselines. Specifically, in the fine-tuning stage of our method, instead of using the proposed GML, we use either balanced cross-entropy or pure cross-entropy but combined with a balanced sampler. All the other settings remain unchanged. The results are shown in~\cref{tab:addition-ablations}. As we can see, our proposed GML is better than them in terms of the harmonic mean of recall and the lowest recall value.

\begin{table}
    \centering
    \resizebox{\columnwidth}{!}{
    \begin{tabular}{lccc}
        \toprule
        Methods&G-Mean&H-Mean&Lowest Recall\\
        \midrule
        CE + GML&\textbf{36.59}&\textbf{31.26}&\textbf{6.00}\\
        \midrule
        CE + CE (re-weighting)&35.30&27.55&4.00\\
        CE + CE (re-sampling)&30.84&18.52&2.00\\
        \bottomrule
    \end{tabular}
    }
    \caption{Comparing with better baselines.}
    \label{tab:addition-ablations}
\end{table}

\subsection{More Visualizations of the Per-Class Recall}

\begin{figure}
    \centering
    \includegraphics[width=0.9\linewidth]{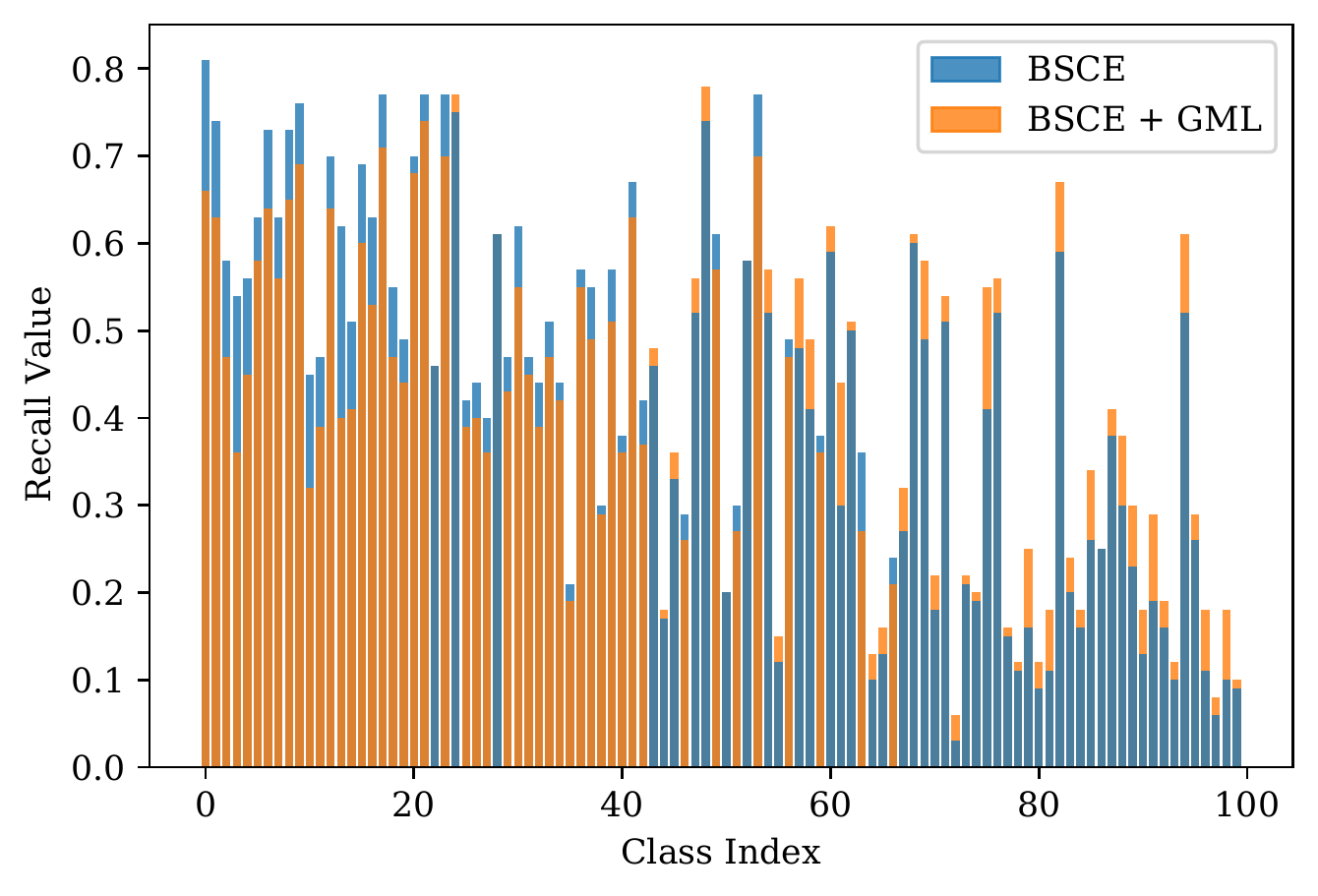}
    \caption{Bar plot of per-class recall on the imbalanced CIFAR100 (with imbalance ratio 100) before and after the fine-tuning when GML is applied to BSCE~\cite{ren2020bsce}.}
    \label{fig:recall-distribution-change-CIFAR100LT-BSCE}
\end{figure}

\begin{figure}
    \centering
    \includegraphics[width=0.9\linewidth]{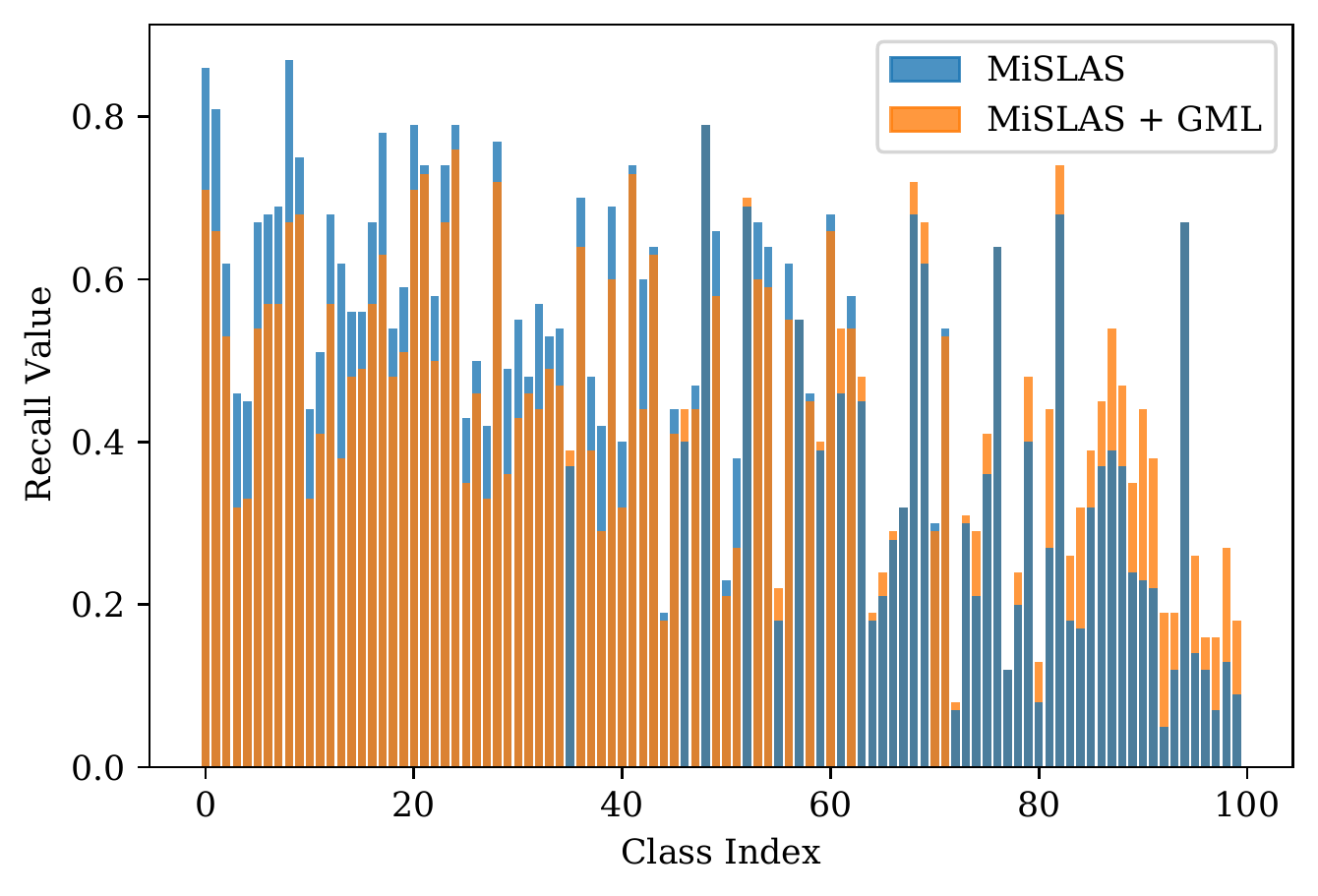}
    \caption{Bar plot of per-class recall on the imbalanced CIFAR100 (with imbalance ratio 100) before and after the fine-tuning when GML is applied to MiSLAS~\cite{zhong2021improving-calibration}.}
    \label{fig:recall-distribution-change-CIFAR100LT-MiSLAS}
\end{figure}

\begin{figure}
    \centering
    \includegraphics[width=0.9\linewidth]{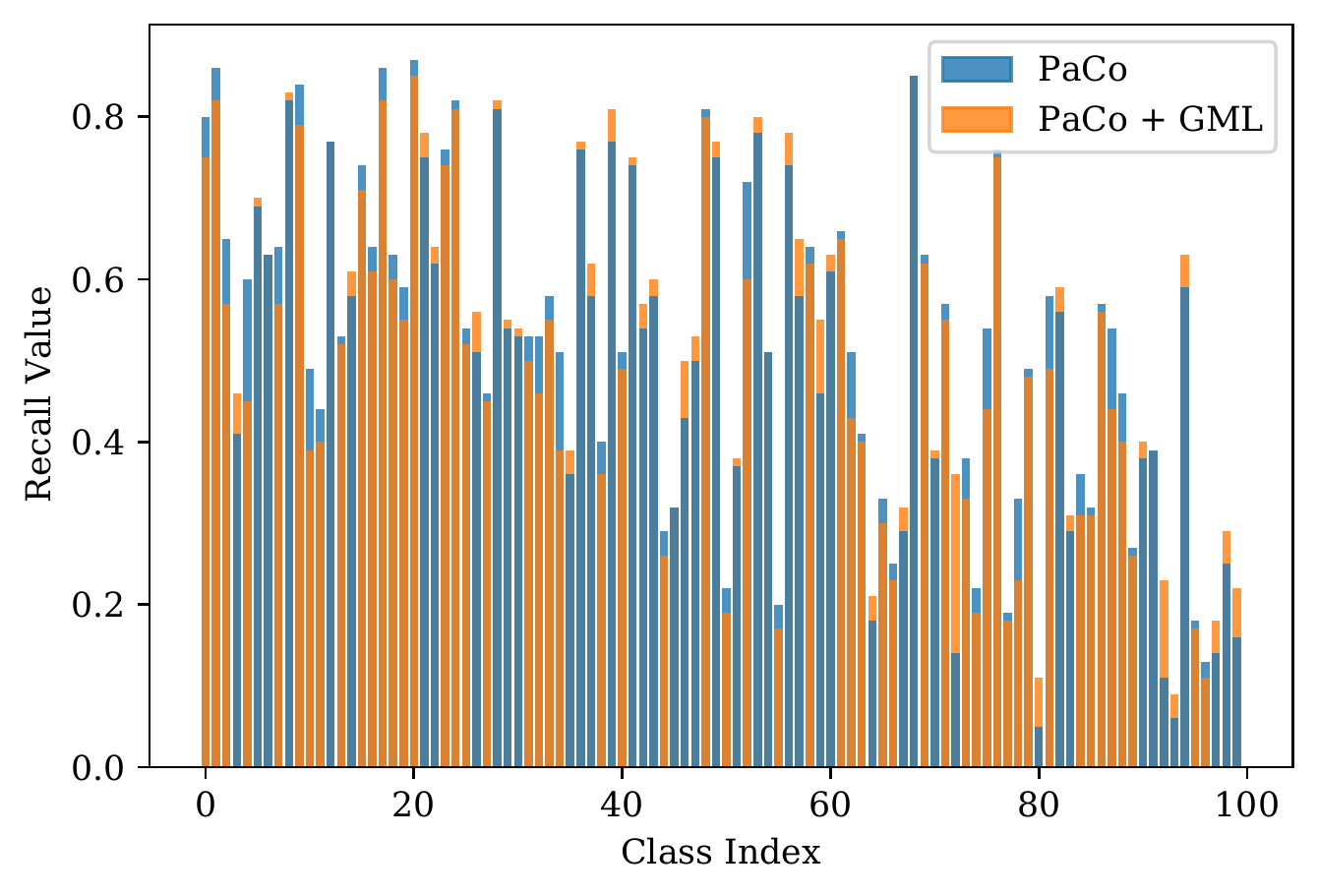}
    \caption{Bar plot of per-class recall on the imbalanced CIFAR100 (with imbalance ratio 100) before and after the fine-tuning when GML is applied to PaCo~\cite{cui2021paco}.}
    \label{fig:recall-distribution-change-CIFAR100LT-PaCo}
\end{figure}

Here we present more visualization results of the per-class recall when GML is applied to different methods. All experiments are conducted on CIFAR100-LT (with imbalance ratio 100). The results are shown in~\cref{fig:recall-distribution-change-CIFAR100LT-BSCE},~\cref{fig:recall-distribution-change-CIFAR100LT-MiSLAS} and~\cref{fig:recall-distribution-change-CIFAR100LT-PaCo}.

\end{document}